%% file: iclr2026_conference.tex
\documentclass{article}
\usepackage{iclr2026_conference,times}
\usepackage{booktabs}
\usepackage{multirow}
\usepackage{makecell}      
\usepackage[table]{xcolor} 
\usepackage{pifont}      
\usepackage{tabularx}
\usepackage{booktabs}
\usepackage{xcolor}
\usepackage[most]{tcolorbox}

\tcbset{
  enhanced,
  enhanced jigsaw,
  breakable,
  fonttitle=\bfseries,
  coltitle=black,
  arc=2mm,
  boxrule=0.6pt,
  left=6pt,right=6pt,top=6pt,bottom=6pt,
  before skip=8pt,after skip=10pt
}

\newtcolorbox{moduleinput}[2][]{
  colframe=#2!70!black,  
  colback=gray!5,         
  colbacktitle=#2!15,
  title={Input},
  #1
}

\newtcolorbox{moduleoutput}[2][]{
  colframe=#2!70!black,   
  colback=yellow!7,       
  colbacktitle=#2!15,
  title={Output},
  #1
}

\tcbuselibrary{listingsutf8}

\definecolor{GreenPigment}{rgb}{0.00,0.65,0.31}
\definecolor{DarkSalmon}{rgb}{0.91,0.59,0.48}
\definecolor{RedOrange}{rgb}{1,0.5,0}
\definecolor{BlueGreen}{rgb}{0.0, 0.5, 0.5}
\usepackage[svgnames]{xcolor}

\newcommand{\blue}[1]{$_{\color{BlueGreen}\downarrow #1}$}
\newcommand{\red}[1]{$_{\color{RedOrange}\uparrow #1}$}

\newcommand{\best}[1]{\textbf{#1}}           
\newcommand{\second}[1]{\underline{#1}}       
\usepackage{array}
\input{math_commands.tex}

\usepackage{amsmath,amssymb,amsfonts,mathtools}
\usepackage{hyperref}
\usepackage{url}
\usepackage{wrapfig}
\usepackage{booktabs}
\usepackage{multirow}
\usepackage{enumitem}

\renewcommand\arraystretch{1.2}
\newcommand{\hdrstrut}{\rule{0pt}{2.2ex}\rule[-0.8ex]{0pt}{0pt}}

\title{{\webweaver}: Dynamic Decomposition and \\ Re-planning for Complex Web Tasks}

\author{Jingbo Yang$^{1}$\thanks{Correspondence to: Jingbo Yang $<$jingbo@ucsb.edu$>$. Work performed while interning at Accenture.}, 
Bairu Hou$^{1}$, 
Wei Wei$^{2}$,
Shiyu Chang$^{1}$,
Yujia Bao$^{2}$ \\
	$^{1}$UC Santa Barbara\quad
    $^{2}$Center for Advanced AI, Accenture \quad
}

\newcommand\webweaver{\textsc{WebDART}}
\iclrfinalcopy 
\begin{document}

\maketitle

\begin{abstract}
Large-language-model (LLM) agents are becoming competent at straightforward web tasks, such as opening an item page or submitting a form, but still struggle with objectives that require long-horizon navigation, large-scale information extraction, and reasoning under constraints. We present {\webweaver}, a general framework that enables a single LLM to handle such complex chores. {\webweaver} (i) \emph{dynamically decomposes} each objective into three focused subtasks—navigation, information extraction, and execution—so the model concentrates on one skill at a time, and (ii) \emph{continuously re-plans} the decomposition as new webpages are revealed, taking advantage of newly discovered filters or shortcuts and avoiding redundant exploration. Evaluated on WebChoreArena, {\webweaver} lifts end-to-end success rates by up to 13.7 percentage points over previous state-of-the-art agents, while matching their performance on the easier WebArena suite and completing tasks with up to 14.7 fewer navigation steps. Code is available at \url{https://github.com/UCSB-NLP-Chang/WebDART}.

\end{abstract}

\section{Introduction}
\label{sec: intro}
\begin{figure}[t]
    \centering
    \includegraphics[width=1.0\textwidth]{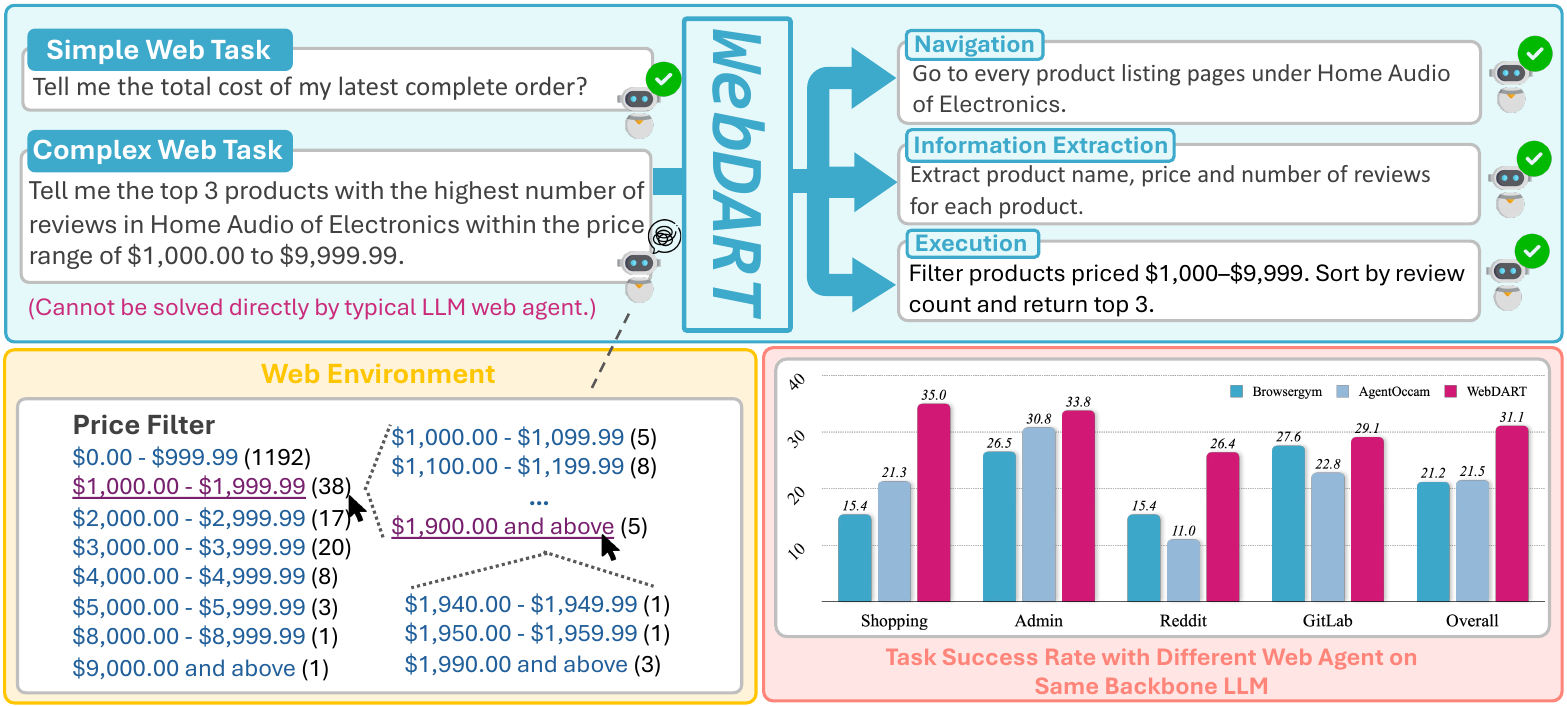} 
    \caption{(Top) Existing LLM-based web agents perform well on simple tasks, but their success rates drop on complex tasks that require non-trivial reasoning, such as applying a price-range filter (bottom left).{\webweaver} overcomes this limitation by dynamically decomposing the objective into three subtasks: navigation, information extraction, and execution. (Bottom right) Consequently,{\webweaver} significantly outperforms the current state of the art on WebChoreArena across all task categories. Backbone LLM: GPT-5.}
    \label{fig: intro}
\end{figure}

LLM-powered web agents have recently shown promising abilities in web navigation tasks~\citep{drouin2024workarena, he2024webvoyager, wei2025browsecomp,yang2024agentoccam,pan2024autonomous,song2024beyond}. Benchmarks such as WebArena~\citep{zhou2023webarena} demonstrate that these agents achieve reasonable accuracy on simple objectives, highlighting their potential as general-purpose automation tools. However, when the objectives require more complex reasoning and multi-step exploration, the performance of these agents often collapses. As shown in Figure~\ref{fig: intro}, on WebChoreArena~\citep{miyai2025webchorearena}, a benchmark designed to test higher-complexity web tasks, agents powered by GPT-4o achieve only 8.0\% accuracy on tasks across different web domains, far below the 46.6\% accuracy on WebArena. This gap highlights a critical weakness of current worflows: while sufficient for simple goals, they are not well equipped for tasks demand multi-step reasoning, long-horizon navigation, and structured information processing.

A closer examination reveals that the difficulty arises from cognitive overload. Complex tasks require agents to simultaneously navigate across multiple web pages, extract and track large amounts of information, and reason under constraints. Consider the following task from WebChoreArena~\citep{miyai2025webchorearena}: \textit{``Tell me the top 3 products with the highest number of reviews in Home Audio of Electronics within the price range of \$1{,}000 to \$9{,}999''}. As illustrated in Figure~\ref{fig: intro}, product information is distributed across multiple nested web pages. Each page may contain tens of products with attributes such as price and number of reviews. To complete this objective, current LLM agents~\citep{yang2024agentoccam, chezelles2024browsergym} attempt to tackle all these aspects in a single process: while browsing through pages, they must also keep track of which products meet the price requirement, remember which ones they have already seen, and simultaneously apply the logic needed to determine the top three by number of reviews. This often overwhelms the agent, leading to frequent mistakes such as missing relevant information, forgetting the user instructions, and incorrect analysis~\citep{miyai2025webchorearena}.

In contrast, human experts may naturally break the task into distinct steps: \ding{182} first narrowing down to the pages within the desired price range, \ding{183} then collecting and recording the attributes of candidate products, and \ding{184} finally ranking the products by number of reviews. This stepwise approach reduces complexity of the task and makes the problem tractable, whereas forcing all operations to occur simultaneously overwhelms current agents and leads to frequent errors.

Motivated by this, we propose \textbf{\webweaver} (\textbf{D}ecomposition \& \textbf{A}daptive \textbf{R}e-planning for \textbf{T}asks), a framework that adaptively decomposes complex web tasks into simpler, modular subtasks. Unlike the typical agentic flow, where navigation, information extraction, and execution are interleaved in a single process, {\webweaver} separates the original complex tasks into these three subtasks. We adopt these three subtasks because complex web tasks typically require distinct agent abilities: browsing through multiple pages, extracting relevant information, and performing analysis or acting on the results.
One example of the decomposition is shown in Figure~\ref{fig: intro}, where we leverage the LLM to generate a decomposition conditioned on both the task description and the initial web environment.
The task decomposition reduces the cognitive burden on the LLM and makes complex objectives more tractable by allowing the agent to focus on one subtask at a time.

However, an initial decomposition based only on the task description may be suboptimal. There are multiple ways to decide what information should be collected during navigation versus deferred to later analysis, and these trade-offs cannot always be known in advance. Moreover, as the agent explores, new web elements such as filters or sort options may appear that were unavailable at the beginning but can drastically reduce navigation effort. For example, in Figure~\ref{fig: intro}, the initial navigation subtask is specified as \emph{“visit every product listing page under Home Audio of Electronics”}. Once the agent enters the product page, it may discover a price filter that allows it to restrict results to \$1{,}000 to \$9{,}999 and avoid traversing irrelevant pages. To exploit such opportunities, {\webweaver} incorporates a \emph{dynamic replanning} mechanism during navigation that allows the agent to revise its plan after each step based on newly observed pages. This adaptive adjustment helps correct mistakes and eliminates redundant exploration.
Together, task-adaptive decomposition and navigation replanning enable {\webweaver} to achieve higher accuracy with lower cost.

We perform extensive evaluation of our method on both WebChoreArena and WebArena across three different LLM backbones. With the proposed decomposition framework, {\webweaver} improves state-of-the-art agent frameworks including BrowserGym~\citep{chezelles2024browsergym} and AgentOccam~\citep{yang2024agentoccam} by up to  13.7\% on the complex tasks in WebChoreArena. Our method also achieves similar performance on WebArena compared to existing state-of-the-arts, demonstrating its robustness and flexibility. Finally, by combining the dynamic re-planning module, the accuracy of our method can be further increased by 7.7\% on the shopping tasks in WebChoreArena while reducing the average navigation steps by 14.7.

\section{Related Work}

\paragraph{Simulated web-agent environments.}
Progress on web agents has largely mirrored progress on the testbeds available to them. The first generation of benchmarks—MiniWoB and MiniWoB++~\citep{shi2017world,liu2018reinforcement}—offers canvas-rendered “toy’’ sites that evaluate low-level actions such as clicking or typing within a single, synthetic page. WebShop keeps the single-domain setting but increases realism by simulating a full e-commerce catalogue, requiring agents to search, filter, and purchase items.

The next wave introduces multi-domain, fully functional sites.
WebArena~\citep{zhou2023webarena} hosts independent applications for shopping, forums, software development, and content management, thereby capturing a broader range of real-world behaviours. More recent suites push two frontiers. (1) Multimodality: VisualWebArena~\citep{koh2024visualwebarena} and WebVoyager~\citep{he2024webvoyager} add image inputs so that agents must reason jointly over text and vision. (2) Task complexity: WebChoreArena~\citep{miyai2025webchorearena} reuses the WebArena sites but issues longer “chores’’ that demand capabilities beyond ordinary browsing—e.g., arithmetic, cross-page memory, and long-horizon planning.

Our study targets the text-only setting and therefore evaluates on WebArena and WebChoreArena, which together provide diverse domains and richly composed task intents while remaining fully reproducible.

\paragraph{LLM-powered web agents.}
Current web agents can be grouped into three broad lines of work.
(1) Leveraging execution feedback. Prompting schemes such as ReAct and its derivatives let an LLM interleave reasoning and actions during a rollout~\citep{yao2023react,mialon2023gaia,hong2024cogagent,yang2024aria,amayuelas2025self,yang2025gta1}. Subsequent methods reuse the generated trajectories to refine future attempts: AWM distils frequently successful action patterns~\citep{wang2024agent}; Auto Eval \& Refine trains an external evaluator and invokes self-reflection~\citep{pan2024autonomous,shinn2023reflexion}; WebPilot explores alternate paths with an MCTS-style search~\citep{zhang2025webpilot}.
(2) Synthesising auxiliary data. Learn-by-Interact creates synthetic tasks, relabels the resulting trajectories with hindsight~\citep{su2025learn,li2020generalized}, and retrieves them at inference time, while AgentSymbiotic uses a large–small model pair to co-generate training examples~\citep{zhang2025symbiotic}. These approaches boost accuracy when the synthetic tasks closely match the evaluation set but risk data contamination and often degrade when distributions diverge.
(3) Optimising the interface. AgentOccam shows that simply pruning the DOM observation and restricting the action set already yields large gains and is now a common preprocessing step~\citep{yang2024agentoccam}.

{\webweaver} departs from all of the above. (i) It is \emph{training-free}: no extra rollouts, synthetic data, or fine-tuning are required. (ii) It tackles long-horizon chores through \emph{dynamic task decomposition}: during execution, the agent continually observes the current webpage and adaptively refines a three-part plan—navigation, information extraction, and execution—allowing the same frozen backbone LLM to focus on one capability at a time. This simple yet principled design delivers state-of-the-art results on both WebArena and WebChoreArena.

\section{Method}
\label{sec: method}
In this paper, we focus on \emph{text-based} web agents, although the proposed approach naturally extends to multimodal environments. Each task is specified by a natural-language instruction and a ground-truth target for evaluation. The agent receives the instruction and interacts with a web environment whose pages are represented as accessibility trees, aiming to fulfil the stated objective.

\begin{figure}[t]
    \centering
    \includegraphics[width=1.0\textwidth]{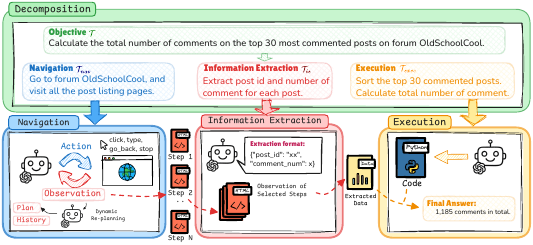} 
    \caption{\textbf{Overview of the \webweaver\ framework.} A complex web task is dynamically decomposed into three sequential subtasks. (1) \textbf{Navigation:} the agent explores the site—issuing actions such as \texttt{click}, \texttt{type}, and \texttt{go\_back}—to gather every page that could contain the required information. (2) \textbf{Information extraction:} given these pages, a dedicated module isolates task-relevant content and converts it into a standardised, structured form based on the objective. (3) \textbf{Execution:} the extracted data are analysed to meet the task constraints, e.g., by generating and running Python code on the fly to perform filtering, aggregation, or other computations.}
    \label{fig: workflow}
\end{figure}

Figure~\ref{fig: workflow} illustrates the \webweaver\ workflow. A complex web task is first \emph{dynamically decomposed} into a sequence of modular subtasks that are executed in order. The central challenge is to choose a decomposition whose subtasks are both tractable and complementary.

Empirically, most web tasks require three distinct capabilities:
\begin{enumerate}[itemsep=0.5em, parsep=0pt]
\item \textbf{Navigation}: browsing across multiple pages to locate candidate information;
\item \textbf{Information extraction}: converting raw page content into structured records;
\item \textbf{Execution}: analysing the collected data or acting on the results.
\end{enumerate}
Guided by this observation, \webweaver\ decomposes every complex task into the ordered subtasks of \emph{navigation}, \emph{information extraction}, and \emph{execution}, continually updating intermediate objectives as new observations arrive. In what follows, we first describe the decomposition strategy (Section~\ref{subsec: task_decomposition}), and then detail the navigation (Section~\ref{subsec: navigation}), information-extraction (Section~\ref{subsec: info_extraction}), and execution (Section~\ref{subsec: exec}) modules.

\subsection{Task Decomposition}
\label{subsec: task_decomposition}

A web task can be decomposed in several ways, and the most suitable granularity depends on the structure of the target site.
Consider the task in Figure~\ref{fig: workflow}:
\emph{``Calculate the total number of comments on the 30 most-commented posts in the OldSchoolCool forum.''}
Two natural decompositions are

\begin{itemize}
    \item \textbf{Tightly coupled.} Embed the numeric constraint in the navigation objective:
          \emph{``Browse OldSchoolCool and open the 30 most-commented posts.''}
    \item \textbf{Conservative.} Keep navigation agnostic to the constraint:
          \emph{``Browse OldSchoolCool and visit all post-listing pages.''}
          Identifying the top 30 posts is then left to the analysis stage.
\end{itemize}

Both options are valid, but their efficiency hinges on site features.
If the forum provides a \texttt{Sort by: most commented} control, the tight plan is ideal—it satisfies the constraint while touching only a handful of pages.
Conversely, when such affordances are absent (or the total number of pages is already small), the conservative plan is simpler and more reliable: the agent just collects every listing page and defers heavy reasoning to later stages.

Because these interface aids are unknown \emph{a priori}, \webweaver\ adopts the conservative scheme by default and adapts opportunistically.  Specifically, all data–centric operations—filtering, sorting, ranking—are initially assigned to execution, while navigation is limited to page discovery.  To steer the LLM toward this partitioning, the prompt $\mathbf{p}$ contains three in-context examples that consistently push constraint handling to later stages:

\[
f : (\mathcal{T}, \mathbf{p}) \;\longrightarrow\;
   \bigl(\mathcal{T}_{\text{nav}},\;
         \mathcal{T}_{\text{ie}},\;
         \mathcal{T}_{\text{exec}}\bigr),
\]

where $f(\cdot)$ is the LLM and the outputs $\mathcal{T}_{\text{nav}}$,     $\mathcal{T}_{\text{ie}}$, $\mathcal{T}_{\text{exec}}$ are the navigation, information-extraction, and execution objectives.

During navigation the agent may encounter helpful widgets (\emph{e.g.}, the aforementioned sort button) that can fulfill part of the constraint immediately.  When detected, \webweaver\ invokes \emph{dynamic replanning}: the current navigation goal $\mathcal{T}_{\text{nav}}$ is updated on-the-fly, allowing the agent to skip irrelevant pages and accelerate completion.  Details of this mechanism are presented in Section~\ref{subsec: navigation}.

\paragraph{Fast-path routing.}
Finally, the decomposition module also incorporates a lightweight router that
decides whether the task can be satisfied with only a \emph{subset} of
the three modules.  For instance, the instruction
``Post \texttt{"Hello, world!"} on \texttt{/OldSchoolCool}" requires
navigation (and possibly execution) but no information extraction; the
router therefore bypasses the extraction stage and invokes the minimal
workflow. 

\begin{wrapfigure}{r}{0.5\linewidth}
  \centering
  \vspace{-4mm}
  \includegraphics[width=\linewidth]{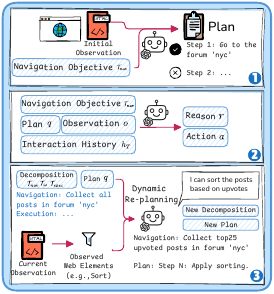}
  \vspace{-8mm}
  \caption{Illustration of the {\webweaver} framework in navigation.
  An initial plan is generated before starting navigation. The navigation agent issues an action at each step. When new web elements (e.g., filters, sorting options) appear, the dynamic re-planning module updates the decomposition and plan, enabling the agent to adapt its strategy for more efficient execution.
  }
  \vspace{-7mm}
  \label{fig:navigation}
\end{wrapfigure}

\subsection{Navigation}
\label{subsec: navigation}

The navigation module drives the agent through the website, issuing low-level browser actions until every page that might contain task-relevant information has been visited.  Our interactive setup follows prior work~\citep{yang2024agentoccam, wang2024agent, zhang2025symbiotic}.

At time step $t$ the agent outputs a pair $(r_t,a_t)$: a natural-language reasoning trace $r_t$ and an action $a_t\!\in\!\mathcal{A}$, where
$\mathcal{A}=\{\texttt{click},\texttt{type},\texttt{go\_back},\texttt{stop}\}$.
The choice is conditioned on
(i) the current navigation objective $\mathcal{T}_{\text{nav}}$,
(ii) the current observation $o_t$ (the page rendered as an accessibility tree), and
(iii) the interaction history
$\mathbf{h}_t=(\mathbf{o}_{1:t-1},\mathbf{a}_{1:t-1},\mathbf{r}_{1:t-1})$.
After execution, $(r_t,a_t)$ is appended to the history; when the agent finally emits \texttt{stop} at step $T$, the full interaction history $\bm h_{T} = (\bm{o}_{[1:T]}, \bm{a}_{[1:T]}, \bm{r}_{[1:T]})$ is passed to the information-extraction module.
Figure \ref{fig:navigation} illustrates the workflow.

\paragraph{Plan-guided browsing.}
Before the first action, the LLM is given the navigation objective $\mathcal{T}_{\text{nav}}$ and the initial page $o_0$ and asked to generate a high-level plan $\mathbf{q}_0$.
The plan lists (i) pages to visit, (ii) information to capture, and (iii) a stopping criterion.
During browsing the agent is prompted with $\mathcal{T}_{\text{nav}}$, the current plan $\mathbf{q}_{t-1}$, the observation $o_t$, and the history $\mathbf{h}_t$.
Conditioning on $\mathbf{q}_{t-1}$ stabilises behaviour and substantially reduces premature termination (sample plans are shown in Appendix \ref{app: navigation}).

\paragraph{Dynamic replanning.}
The conservative decomposition from Section \ref{subsec: task_decomposition} defers all constraint handling to the execution stage; this guarantees coverage but can be wasteful when helpful interface widgets (filters, sort menus, etc.) appear mid-navigation.  To exploit such shortcuts, the agent performs \emph{dynamic replanning}.

At the start of each step $t$ the agent evaluates, based on $(o_t,\mathbf{h}_{t-1},\mathbf{q}_{t-1},\mathcal{T})$, whether the navigation objective or plan should be revised.
If a useful widget has been discovered, it outputs an updated pair $(\mathcal{T}_{\text{nav}}^{\,t},\mathbf{q}_t)$ that incorporates the shortcut; otherwise it keeps the previous version.
The (possibly) updated objective and plan are fed back into the action-selection prompt to produce $(r_t,a_t)$.

Dynamic replanning preserves the safety of a conservative start while allowing the agent to exploit opportunistic efficiencies—for example, switching from “visit every listing page’’ to “apply \texttt{sort by: most-commented} and scan only the first 30 posts.’’  The prompt template used for this mechanism is provided in Appendix \ref{app: re-planning}.

\subsection{Information Extraction}
\label{subsec: info_extraction}
When navigation ends at step $T$, we obtain the transcript $\mathbf{h}_T = (\mathbf{o}_{1:T}, \mathbf{a}_{1:T}, \mathbf{r}_{1:T})$, where $\mathbf{o}_{1:T}$ contains every page the agent observed. Blindly extracting from \emph{all} pages would add substantial noise—for example, products in the wrong category or outside a specified price range.  The extraction module therefore proceeds in two stages:

\paragraph{Page selection.}
An LLM is given the original task $\mathcal{T}$, the navigation objective $\mathcal{T}_{\text{nav}}$, and the full history $\mathbf{h}_T$.  It returns an index set $\mathcal{I}\subseteq\{1,\ldots,T\}$ that marks the pages most likely to contain the required information (prompt template in Appendix~\ref{app: information extraction}).

\paragraph{Field extraction.} For each chosen page $o_t\;(t\!\in\!\mathcal{I})$, a second LLM call extracts the target fields—\emph{e.g.}, post title and comment count—directly from the page’s accessibility tree, producing a uniform JSONL record. The resulting structured collection is passed to the execution module.

We also experimented with an \emph{LLM-generated parser} baseline, where the model generates code on the fly to traverse the accessibility tree of each $o_t$.  In practice, this approach proved brittle: accessibility trees are deeply nested and site-specific, and minor layout changes frequently break the generated code.  Prompt-based extraction avoids these issues and requires no hand-crafted logic; therefore, \textsc{WebWeaver} adopts it as the default strategy.

\subsection{Execution}
\label{subsec: exec}

The execution module converts the structured records produced by the
information‐extraction stage into the final deliverable requested by the
task.  Depending on $\mathcal{T}_{\text{exec}}$, this entails one of two
sub-routines.

\paragraph{Data-analysis objectives.}
When the task calls for statistics, rankings, or other derived
quantities, the agent generates and runs code (Python by default) over
the extracted JSON file.  Typical operations include filtering under
constraints, aggregation, and sorting.  To increase robustness we adopt
a \emph{self-reflection} loop \citep{shinn2023reflexion}: if the program
throws an exception, the LLM examines the traceback, amends the code,
and re-executes it until success or a timeout.  Implementation details
are provided in Appendix~\ref{app: details}.

\paragraph{Action-oriented objectives.}
Some tasks require injecting the computed result back into the
environment—for example, posting a summary to a forum or submitting a
completed form.  In these cases the module invokes a short-horizon
navigation policy that is initialised with the analysis output (e.g.,
the text to post or the value to enter).  Because the destination
elements are already known, this policy is far simpler and more reliable
than the primary navigation module, yet it preserves the same interface
and action space.

In both settings, once the required code or interactions have concluded,
the agent returns the task’s final answer and the execution stage
terminates.

\section{Experiment Results and Analysis}

\subsection{Experiment Setup}
\label{subsec: exp setup}
\paragraph{Environment.} We conduct experiments on two benchmarks: \textbf{WebChoreArena} and \textbf{WebArena}. WebChoreArena~\citep{miyai2025webchorearena} is our primary evaluation benchmark, as it extends the WebArena~\citep{zhou2023webarena} environment with more realistic and challenging chores that require handling constraints, information extraction, and data analysis in addition to navigation. These tasks better reflect the complexity of real-world web usage and thus serve as the main testbed for demonstrating the effectiveness of our method. In parallel, we also evaluate on WebArena tasks to ensure that our approach does not reduce performance on simpler navigation-oriented objectives. Both benchmarks share the same set of interactive web environments (e.g., shopping, administration, forums, and code management), which allows us to make a direct comparison between simple and complex tasks under consistent conditions.

\paragraph{Baselines.} 
We compare {\webweaver} against four baselines: \textbf{SteP}~\citep{sodhi2023step}, \textbf{BrowserGym}~\citep{chezelles2024browsergym}, \textbf{AWM}~\citep{wang2024agent} and \textbf{AgentOccam}. SteP~\citep{sodhi2023step} (Stacked LLM Policies) is a method that decomposes the web‐agent policy space into multiple sub‐policies, dynamically composing them to adapt to task complexity. BrowserGym~\citep{chezelles2024browsergym} provides a unified evaluation framework for web agents with standardized observation and action spaces, enabling fair and reproducible comparisons across different benchmarks. AWM~\citep{wang2024agent} induce commonly reused rountines from web tasks to guide subsequent generations.
AgentOccam~\citep{yang2024agentoccam} is our main baseline, as it employs a navigation agent design closely aligned with ours; by focusing on observation and action spaces that match LLM pretraining distributions, it achieves strong results on WebArena without relying on in‐context examples or external search. Together, these baselines allow us to evaluate {\webweaver} against diverse approaches while ensuring a fair comparison with a closely related navigation agent. We compare {\webweaver} with these baselines with three different backbone LLMs including GPT-5, GPT-4o, and GLM-4.5-air-fp8. The configurations for each model and experiment setup is detailed in Appendix~\ref{app: details}

\subsection{Evaluation on complex web tasks.}

\input{tables/main}

Table~\ref{tab: main} presents the main results on the \textbf{WebChoreArena} benchmark, which evaluates agent performance on complex multi-step web tasks involving constraints and information extraction. We compare {\webweaver} against three baselines: SteP, AWM, BrowserGym, and AgentOccam, under three different backbone models (GPT-5, GPT-4o, and GLM-4.5-air-fp8). 

Across all model backbones, {\webweaver} achieves the highest overall success rates, demonstrating its robustness and effectiveness on complex tasks. With GPT-5, {\webweaver} reaches 31.1 overall, outperforming SteP (3.1), BrowserGym (21.2), AWM (22.4), and AgentOccam (21.5). The gains are particularly pronounced in the Shopping and Reddit domains, where {\webweaver} improves over AgentOccam by +13.7 and +15.4 points respectively. This highlights the advantage of shifting constraint handling to the data analysis stage, which reduces error propagation from fragile navigation.

The improvements are consistent for GPT-4o, where {\webweaver} achieves 15.2 overall compared to 8.0 for AgentOccam, and for GLM-4.5-air-fp8, where {\webweaver} reaches 19.3 overall compared to 10.8 for AgentOccam. These results suggest that our method generalizes across different backbone models, even when the underlying LLM has weaker navigation or reasoning capabilities. 

We also note that SteP underperforms significantly on WebChoreArena compared to other baselines and {\webweaver}, reflecting its limited ability to handle tasks with deep constraint hierarchies. In contrast, {\webweaver} consistently maintains a strong margin over all baselines, confirming that decomposition is the key to solving complex web chores efficiently.

\subsection{Evaluation of dynamic re-planning.}

In Section~\ref{subsec: navigation}, we introduced \textit{dynamic re-planning}, where the navigation agent adapts its decomposed subtasks and plan based on newly discovered web elements (e.g., filters or sorting options) that can directly apply task constraints. This mechanism aims to reduce redundant navigation and improve efficiency, while preserving or even improving accuracy. Table~\ref{tab:efficiency} reports the results of comparing agents with and without dynamic re-planning across four domains in  using GPT-4o as the backbone model. We report both task accuracy and the average number of navigation steps. 

The results show that dynamic re-planning substantially reduces the number of navigation steps. In the Shopping domain, the average navigation steps decrease from 32.9 to 18.2 while accuracy improves from 18.8\% to 26.5\%. A similar trend is observed in Reddit, where the step count drops from 25.1 to 20.8, with a modest accuracy gain (19.8\% to 20.9\%). The only exception occurs in the Shopping Admin domain. This is because the website inherently relies on numerous filters and sorting elements, without which the tasks cannot be completed. These improvements confirm that dynamically adapting the decomposition and plan allows the agent to bypass unnecessary exploration and focus on relevant parts of the environment. 

\input{tables/efficiency}

\input{tables/webarena}
Overall, these results validate the effectiveness of dynamic re-planning as a complementary strategy in {\webweaver}. By allowing the agent to adjust its task structure in real time, we achieve shorter navigation paths and, in several domains, notable accuracy improvements. 

\subsection{Evaluation on simple navigation tasks.}

While {\webweaver} is primarily designed for complex web tasks involving constraints and analysis, it is also important to verify that the framework does not degrade performance on simpler navigation-oriented tasks. To this end, we evaluate on the original \textbf{WebArena} benchmark, where most tasks can be completed through direct navigation without requiring decomposition. For these tasks, we adjust the agent to bypass the decomposition stage and focus solely on the navigation module. 

Table~\ref{tab:webarena} reports the results, comparing {\webweaver} against a wide range of existing web agents. We observe that {\webweaver} achieves competitive or superior performance across domains, reaching an overall success rate of 48.1, which is higher than all baselines including AgentOccam (46.6).

These results confirm that {\webweaver} maintains robustness across task types: it significantly improves over baselines in complex settings by leveraging decomposition, while also remaining competitive on simpler navigation tasks by bypassing unnecessary modules. This adaptability demonstrates the generality of our design.

\subsection{Case study.}
We further present case study to visualize how dynamic re-planning enhances {\webweaver} in Table~\ref{tab:replanning-cases}.
In the first example, the agent initially plans to traverse every page in a product category, but upon detecting a drop-down menu that adjusts the number of displayed products, the plan is revised to greatly reduce navigation steps. This shows how re-planning exploits newly discovered web elements to improve efficiency. In the second case, the agent’s initial decomposition requires visiting all forums to collect a user’s submissions, which is infeasible. Once it identifies that the user profile page already lists submissions with a direct link, the plan and the navigation objective is updated to extract information more directly, correcting a flawed decomposition. Finally, in the third case, the agent relies on keyword search that produces irrelevant results. Dynamic re-planning detects the mismatch and redirects the strategy to the actual forum page, enabling the agent to recover from misleading navigation. Together, these examples demonstrate that dynamic re-planning allows the agent to correct initial mistakes and maintain robustness in complex web environments.

\input{tables/case}

\section{Conclusion}

We introduced {\webweaver}, a framework that enhances web agents on complex tasks through explicit subtask decoupling and dynamic re-planning. By shifting constraint handling and other data-related operations from navigation to the analysis stage, {\webweaver} reduces error propagation and alleviates the burden on fragile navigation processes. At the same time, dynamic re-planning enables the agent to adapt plans in real time when new web elements are discovered or when the initial decomposition is suboptimal. Experiments on WebChoreArena demonstrate that {\webweaver} improves task success rates by up to 13.7\% over strong baselines while also reducing navigation steps, and evaluation on WebArena confirms that our method maintains performance on simpler tasks. Case studies further show how re-planning allows the agent to exploit new opportunities, correct inefficient strategies, and recover from misleading navigation paths, leading to more efficient and robust web automation.

\section*{Acknowledgment}
The work of Bairu Hou and Shiyu Chang was partially supported by National Science
Foundation (NSF) Grant IIS-2338252, NSF Grant IIS-2207052, and NSF Grant IIS-2302730.

\bibliography{iclr2026_conference}
\bibliographystyle{iclr2026_conference}

\newpage
\appendix
\section{Appendix}
\label{app: prompt}
\subsection{Agent Prompts \& Examples}
Inside this section, we displayed the prompts as well as some intermediate outputs as demonstration examples for for each module of {\webweaver}.

\subsubsection{Decomposition}
\label{app: decomposition}
The following prompt illustrates an example of decomposition for data-analysis objectives. It explicitly encourages a conservative strategy, as discussed in our method section, by deferring data-related operations to the analysis stage. In addition, we provide three in-context examples to help the LLM better follow this decomposition approach.

\begin{tcolorbox}[enhanced jigsaw, breakable, title=Prompt - Decomposition, colback=gray!3, colframe=green!50]
You are conducting a complex web task that requires information from the web to answer correctly. Directly navigating the web environment to provide a final answer cannot always yield the correct result. Therefore, you need to decompose the task into two decoupled parts to complete it successfully.

\vspace{1em}
The two parts are the navigation part and the analysis part.
The navigation part involves visiting all pages that contain the data needed to solve the task. The observation, the accessibility tree of full web page, at each step will be recorded during navigation.

The analysis part involves extracting information from the observations and writing code to provide the final answer. Note that the extracted information processed during analysis part may be imperfect, which means they may include unnecessary data or not in correct format, you need to make sure the analysis code can be robust to handle such cases.

\vspace{1em}
Another important consideration is to simplify the navigation, as it is a more challenging task. Ignore constraints such as ranges or filters in the navigation objective. Instead, include such constraints in the analysis part to be handled later.

\vspace{1em}
Given the original complex user task and some tips for using the target website, decompose it into these two parts following this approach. Your output must follow this format with exact the same headers:

\vspace{1em}
\textbf{\#\#\# Part 1 – Navigation}

\vspace{1em}
\textbf{\#\#\# Part 2 – Analysis}

\vspace{1em}
In addition, below are some decomposition examples for your reference:

\vspace{1em}
\textbf{Example 1:}

\vspace{1em}
User task  
“List the average rating for every movie genre, using only titles released between 2015 and 2024. Output: ‘Drama : 8.1, Comedy : 7.4, …’”

\vspace{1em}
\#\#\# Part 1 – Navigation  
Go to the pages which include each film’s genre, release year, and numeric user rating. Do not go to each film detail page if all the information is available in film listing page.

\vspace{1em}
\#\#\# Part 2 – Analysis  
Filter and only keep only films released 2015-2024. Compute the average rating per genre and show them as ‘Drama : X.X, Comedy : Y.Y, …’.

\vspace{1em}
\textbf{Example 2:}

\vspace{1em}
User task  
“Among products tagged ‘wireless earbuds’, count how many cost below \$50, \$50-\$99, and \$100+. Return: ‘$<$50 : \_\_, 50-99 : \_\_, 100+ : \_\_’.”

\vspace{1em}
\#\#\# Part 1 – Navigation  
Visit the pages containing product title and price information for “wireless earbuds” products. Do not go to each product detail page if all the information is available in product listing page.

\vspace{1em}
\#\#\# Part 2 – Analysis  
Group the collected items by price brackets  $<$ \$50, \$50-\$99, \$100+.  Count how many fall into each bracket and output the counts in the following format: ‘$<$50 : \_\_, 50-99 : \_\_, 100+ : \_\_’

\vspace{1em}
\textbf{Example 3:}

\vspace{1em}
User task  
“In the travel forum, among the 200 latest hotel reviews, how many mention ‘noise’ or ‘quiet’ in the text? Give two numbers: noisy\_count, quiet\_count.”

\vspace{1em}
\#\#\# Part 1 – Navigation  
Navigate to the pages including the text body of the hotel reviews in most recent order in the travel forum. Go over all hotel reviews in total. Do not go to each review detail page if all the information is available in review listing page.

\vspace{1em}
\#\#\# Part 2 – Analysis  
Only keep first 200 reviews. Search each saved review for the words “noise”, “noisy” (noisy\_count) and “quiet”. Return two integers: noisy\_count and quiet\_count.

\end{tcolorbox}

Below is one decomposition example generated conditioned on the prompt above:

\begin{tcolorbox}[enhanced jigsaw, breakable, title=Example - Decomposition, colback=gray!8, colframe=green!50]
\textbf{Original Task:} 

\vspace{1em}
Extract the title of reviews with a rating of 2 or below out of 5 stars from `Tea Gift Set for Tea Lovers - Includes Double Insulated Tea Cup 12 Uniquely Blended Teas and All Natural Honey Straws | Tea Gift Sets for Women Men | Tea Gifts Bag Presented in Beautiful Gift Bag' and output them as a list in alphabetical order, separeted by line breaks.

\vspace{1em}
\textbf{Navigation Objective:} 

\vspace{1em}
Navigate to the product page for `Tea Gift Set for Tea Lovers - Includes Double Insulated Tea Cup 12 Uniquely Blended Teas and All Natural Honey Straws | Tea Gift Sets for Women Men | Tea Gifts Bag Presented in Beautiful Gift Bag'. Visit the reviews section of the product and collect the review titles along with their star ratings.

\vspace{1em}
\textbf{Analysis Objective:} 

\vspace{1em}
Filter the collected reviews to include only those with a rating of 2 stars or below. Extract the titles of these reviews and sort them in alphabetical order. Output the sorted titles as a list, with each title separated by a line break.

\end{tcolorbox}

\subsubsection{Navigation}
\label{app: navigation}

In this section, we display the prompts for each part of navigation module and provide corresponding examples.

\begin{tcolorbox}[enhanced jigsaw, breakable, title=Prompt - Navigation, colback=gray!3, colframe=blue!50]
You are an AI assistant performing navigation tasks on a web browser. You will be provided with task objective, current step, web page observations, current plan, and interaction history. You need to issue an action for this step.

\vspace{1em}
Your task is mainly about navigating to each page that may contain the needed information.

\vspace{1em}
Generate the response in the following format:
\{output\_specifications\}

\vspace{1em}
You are ONLY allowed to use the following action commands. Strictly adheres to the given format. Only issue one single action.
\{navigation\_specifications\}

\vspace{1em}
\{website\_tips\}
\end{tcolorbox}

\begin{tcolorbox}[enhanced jigsaw, breakable, title=Prompt - Navigation Plan Generation, colback=gray!3, colframe=blue!50]

You are an AI assistant that generates initial plans for web navigation tasks. Given a task objective and an initial web page observation in accessibility tree, you need to create a clear, step-by-step plan that will guide the navigation agent.

\vspace{1em}
Directly output the navigation plan in your response without other irrelevant information.

\vspace{1em}
Your plan should be:

\vspace{1em}
1. Clear and actionable

\vspace{1em}
2. Broken down into logical steps

\vspace{1em}
3. Specific enough to guide navigation

\vspace{1em}
4. Concise, only including necessary steps. Do not dive into more pages if the current page already contains the needed information

\vspace{1em}
5. Focused only on how to navigate, do not include other steps including extraction, analysis, opening website, closing environment, etc

\vspace{1em}
Consider common web navigation patterns like:

\vspace{1em}
- Searching for information

\vspace{1em}
- Navigating through menus and links

\vspace{1em}
- Going over the necessary pages

\vspace{1em}
- Interacting with buttons and controls

\end{tcolorbox}

\begin{tcolorbox}[enhanced jigsaw, breakable, title=Example - Navigation Plan, colback=gray!8, colframe=blue!50]
\textbf{Objective: }

\vspace{1em}
Search for the product ``Manhattan Comfort Avesta Double Side Table 2.0 Collection Free Standing Modern Side Table / TV Stand with Storage Includes 2 Doors with 3 Shelves and Features Splayed Legs, White/Stamp/Grey Legs'' on the website. Navigate to the product page that contains detailed information including customer reviews. Ensure to collect review data that includes the rating scores or star ratings.

\vspace{1em}
\textbf{Plan:}

\vspace{1em}
1. Locate the search combobox labeled \"Search\" in the accessibility tree.

\vspace{1em}
2. Input the product name: ``Manhattan Comfort Avesta Double Side Table 2.0 Collection Free Standing Modern Side Table / TV Stand with Storage Includes 2 Doors with 3 Shelves and Features Splayed Legs, White/Stamp/Grey Legs\'' into the search combobox.

\vspace{1em}
3. Locate and activate the ``Search'' button to initiate the product search.

\vspace{1em}
4. On the search results page, look for the product name or a closely matching link.

\vspace{1em}
5. Click on the link corresponding to the desired product to navigate to its detailed product page.

\vspace{1em}
6. On the product page, locate the section containing customer reviews or star ratings.
\end{tcolorbox}

\subsubsection{Information Extraction}
\label{app: information extraction}

During information extraction, we first instruct the LLM to select relevant observations based on the overall trajectory. The prompt is given as follows.
\begin{tcolorbox}[enhanced jigsaw, breakable, title=Prompt - Relevant Observation Selection, colback=gray!3, colframe=red!30]
        You are a judge agent in a web navigation and information seeking task.

\vspace{1em}
        Given a navigation objective (which includes the information to be found in the web environment) and a list of web navigation agent interaction history (with reason, action, and observation summary), 
        select the step numbers that their observations are most likely to contain the information specified in the objective.

\vspace{1em}
        Analyze each step in one or two sentences. After this, return a JSON list of step numbers (e.g., [2, 5, 7]) that you believe contains the needed information in their observations.
        Note:
        
\vspace{1em}
        1) The action in a step will be executed and reflected in the observation in the next step. For example, if the action is `click on the home page button', the observation in the next step will be the home page.

\vspace{1em}
        2) The action you see at each step may contain a number, like `click[1316]'. This number is the index of the element in the observation. You may not know which element is clicked, but you can still use the reason to infer what that element is.
\end{tcolorbox}

After selecting the relevant observations, we will first let the LLM to generate a prompt for extraction at each page. The reason for this step is to fix a data schema for easily integrating results from multiple pages.
\begin{tcolorbox}[enhanced jigsaw, breakable, title=Prompt - Extraction Prompt Engineering, colback=gray!3, colframe=red!30]
        You are an expert prompt engineer.
        Design a SINGLE prompt that, when shown together with a web-page 
        text accessibility tree, makes another LLM extract and return ONLY a list of JSON object
        containing the fields that satisfy the user goal.
        Only extract the information specified in the user goal. Make sure each extracted entry also has one identifier field (add only one if there is no such key specified in user goal) that will helps accurate deduplication in the later stage.
        You need to specify 1) what information to be extracted, 2) what keys should be used for each JSON object in extracted list, 3) one simple example of the extracted JSON list.
        Make your prompt concise and only include these necessary infromation.
\end{tcolorbox}

\subsubsection{Execution}
\label{app: execution}

Below we provide the prompt for writing data analytic code during execution phase.
\begin{tcolorbox}[enhanced jigsaw, breakable, title=Prompt - Data Analysis, colback=gray!3, colframe=Yellow!30]
        You are an analysis assistant that MUST write Python code.

        \vspace{1em}
        You will be provided with objective and data samples (a small portion of all the data as a reference) for analysis as a reference.

        \vspace{1em}
        • The data is pre-loaded in a variable named \verb|`|data\verb|`|.

        \vspace{1em}
        • Assign your final answer to a variable named \verb|`|answer\verb|`|.

        \vspace{1em}
        Return only one fenced block:

        \vspace{1em}
        \verb|```|python\# code here
        
        \vspace{1em}
        answer = ...\verb|```|
\end{tcolorbox}

\subsubsection{Re-planning}
\label{app: re-planning}

We provide the prompt of re-planning and one example here.

\begin{tcolorbox}[enhanced jigsaw, breakable, title=Prompt - Re-planning, colback=gray!3, colframe=purple!30]

You are a Dynamic Control Agent responsible for monitoring and adapting the task decomposition and navigation plan based on new observations during web navigation.

\vspace{1em}
Your role is to:

1. Assess whether the current decomposition and navigation plan are still appropriate given the new web elements and information discovered

2. Determine if modifications are needed to better achieve the original objective

3. Update the decomposition and navigation plan when necessary

\vspace{1em}
You will be provided with:
- The original task objective
- Current decomposition (Part 1 - Navigation, Part 2 - Analysis)
- Current navigation plan
- Current web page observation
- Interaction history

\vspace{1em}
Based on this information, you need to decide whether to:

- Keep the current decomposition and navigation plan unchanged

- Modify the decomposition to better reflect what needs to be done

- Update the navigation plan to account for new web elements or information discovered

\vspace{1em}
Adhere to the following output format:
\{output\_specifications\}

\vspace{1em}
Guidelines:

- Only modify decomposition/plan if you discover new web elements or information that significantly changes the approach

- Be conservative - don't change things unnecessarily

- Focus on practical improvements that will help achieve the objective more effectively

- Consider if new navigation paths or information sources have been discovered

- Ensure any updates are clear and actionable

- Do not include any task in analysis objective into the plan, as the plan is only for navigation

\vspace{1em}
\{website\_tips\}

\end{tcolorbox}

\begin{tcolorbox}[enhanced jigsaw, breakable, title=Example - Re-planning, colback=gray!8, colframe=purple!30]

\textbf{Initial navigation plan}:

\vspace{1em}
1. Navigate to the "Forum" tab on the website.

\vspace{1em}
2. Click on "alphabetical" to view all forums listed alphabetically.

\vspace{1em}
3. Visit each forum and locate submissions made by the user "thebelsnickle1991".

\vspace{1em}
4. If there are multiple pages for submissions in a forum, continue navigating using the "More" link until reaching the last page.

\vspace{1em}
\textbf{Re-planning reasoning:}

\vspace{1em}
The current observation reveals that we are already on a page with submissions by the user ``thebelsnickle1991.'' However, the original objective and decomposition specify starting from the ``Forum'' tab, then navigating to an alphabetical list of forums, and finally visiting pages containing submission data for this user. The current page already provides access to numerous submissions by the user, categorized under different forums, with a "More" link at the bottom for additional pages. This suggests a more direct path to fulfilling the objective.  

The decomposition and navigation plan should be updated to prioritize directly extracting and analyzing the user's submissions from this page and subsequent pages accessed via the ``More'' link. This eliminates the need to navigate to the ``Forum'' tab or the alphabetical list of forums, streamlining the process.

\vspace{1em}
\textbf{Updated navigation plan:}

\vspace{1em}
1. Extract submission data for "thebelsnickle1991" directly from the user page.  

\vspace{1em}
2. Use the "More" link to navigate through additional pages containing submissions by "thebelsnickle1991" and extract data from those pages.

\end{tcolorbox}

\subsubsection{Others}
Here we provide the prompt detail of the website tips we used and navigation specification for the navigation prompts above.

Following the WebChoreArena~\citep{miyai2025webchorearena}, we used website tips for the evaluation in our experiments for our method and all the other baselines. 
\begin{tcolorbox}[enhanced jigsaw, breakable, title=Prompt - Website Tips, colback=gray!3, colframe=pink!50]
\begin{tcolorbox}[enhanced jigsaw, breakable, title=Shopping,coltitle=black, colback=gray!5, colframe=pink!50]
1. This website provides very detailed category of products. You can hover categories on the top menu to see subcategories.

\vspace{1em}
2. If you need to find information about your previous purchases, you can go My Account $>$ My Orders, and find order by date, order number, or any other available information

\vspace{1em}
3. An order is considered out of delivery if it is marked as "processing" in the order status

\vspace{1em}
4. When the task asks you to draft and email. DO NOT send the email. Just draft it and provide the content in the last message

\vspace{1em}
5. If the review star rating is not directly available but the rating score is provided, you can estimate the star rating by dividing the rating score by 20. For example, a rating score of 80 corresponds to a 4-star review

\vspace{1em}
6. Utilize the search if you need to find the information of a specific item, and use the top menu when you need to visit a category
\end{tcolorbox}

\begin{tcolorbox}[enhanced jigsaw, breakable, title=Shopping Admin,coltitle=black, colback=gray!5, colframe=pink!50]
Here are tips for using this website:

\vspace{1em}
1. When you add a new product in the CATALOG $>$ Products tab, you can click the downwardarrow beside the "Add Product" button to select options like "Simple Product", "Configurable Product", etc.

\vspace{1em}
2. If you need to add new attribute values (e.g. size, color, etc) to a product, you can find the product at CATALOG $>$ Products, search for the product, edit product with "Configurable Product" type, and use "Edit Configurations" to add the product with new attribute values. If the value that you want does not exist, you may need to add new values to the attribute.

\vspace{1em}
3. If you need to add new values to product attributes (e.g. size, color, etc), you can visit STORES $>$ Attributes $>$ Product, find the attribute and click, and add value after clicking "Add Swatch" button.

\vspace{1em}
4. You can generate various reports by using menus in the REPORTS tab. Select REPORTS $>$ "report type", select options, and click "Show Report" to view report.

\vspace{1em}
5. In this website, there is a UI that looks like a dropdown, but is just a 1-of-n selection menu. For example in REPORTS $>$ Orders, if you select "Specified" Order Status, you will choose one from many options (e.g. Canceled, Closed, ...), but it's not dropdown, so your click will just highlight your selection (1-of-n select UI will not disappear).

\vspace{1em}
6. Configurable products have some options that you can mark as "on" of "off". For example, the options may include "new", "sale", "eco collection", etc.

\vspace{1em}
7. You can find all reviews and their counts in the store in MARKETING $>$ User Content $>$ All Reviews. If you see all reviews grouped by product, go REPORTS $>$ By Products and search by Product name.

\vspace{1em}
8. This website has been operating since 2022. So if you have to find a report for the entire history, you can select the date from Jan 1, 2022, to Today.

\vspace{1em}
9. Do not export or download files, or try to open files. It will not work.
\end{tcolorbox}

\begin{tcolorbox}[enhanced jigsaw, breakable, title=Reddit,coltitle=black, colback=gray!5, colframe=pink!50]
Here are tips for using this website:

\vspace{1em}
1. when the task mentions subreddit, it is referring to ‘forum'

\vspace{1em}
2. if you need find a relevant subreddit or forum, you can find the name after clicking "alphabetical" in the "Forum" tab

\vspace{1em}
3. you can visit the next page with the link 'More', if the link 'More' is NOT visible in the current observation, this means you have reached the last page
\end{tcolorbox}

\begin{tcolorbox}[enhanced jigsaw, breakable, title=Gitlab,coltitle=black, colback=gray!5, colframe=pink!50]
1. your user name is byteblaze

\vspace{1em}
2. To add new members to the project, you can visit project information $>$ members tab and click blue "invite members" button on top right

\vspace{1em}
3. To set your status, click profile button on top right corner of the page (it's next to the question mark button) and click edit status

\vspace{1em}
4. To edit your profile, click profile button on top right corner of the page (it's next to the question mark button) and click edit profile

\vspace{1em}
5. You can also access to your information e.g. access token, notifications, ssh keys and more from "edit profile" page

\vspace{1em}
6. Projects that you have contributed to are listed under Project / Yours / All tab of gitlab.site. You can sort repos using dropdown button on top right

\vspace{1em}
7. Projects's repository tab has menus like Commits, Branches, Contributors, and more. Contributors tab shows contributors and their number of commits

\vspace{1em}
8. If you want to see all the issues for you, you can either click button on the right of + icon on top right menu bar

\vspace{1em}
9. When the task mentions branch main, it often means master
\end{tcolorbox}
\end{tcolorbox}

\begin{tcolorbox}[enhanced jigsaw, breakable, title=Prompt - Navigation Specification, colback=gray!3, colframe=pink!50]
\begin{tcolorbox}[enhanced jigsaw, breakable, title=``click'',coltitle=black, colback=gray!5, colframe=pink!50]
click [id]: To click on an element with its numerical ID on the webpage. E.g., `click [7]` If clicking on a specific element doesn't trigger the transition to your desired web state, this is due to the element's lack of interactivity or GUI visibility. In such cases, move on to interact with OTHER similar or relevant elements INSTEAD.
\end{tcolorbox}
\begin{tcolorbox}[enhanced jigsaw, breakable, title=``go\_back'',coltitle=black, colback=gray!5, colframe=pink!50]
go\_back: To return to the previously viewed page.
\end{tcolorbox}
\begin{tcolorbox}[enhanced jigsaw, breakable, title=``type'',coltitle=black, colback=gray!5, colframe=pink!50]
type [id] [content] [press\_enter\_after=0/1]: To type content into a field with a specific ID. By default, the "Enter" key is pressed after typing unless \verb|`|press\_enter\_after\verb|`| is set to 0. E.g., \verb|`|type [15] [Carnegie Mellon University] [1]\verb|`| If you can't find what you're looking for on your first attempt, consider refining your search keywords by breaking them down or trying related terms.
\end{tcolorbox}
\begin{tcolorbox}[enhanced jigsaw, breakable, title=``stop'',coltitle=black, colback=gray!5, colframe=pink!50]
stop [answer]: To stop interaction and return response. ONLY use this action when you believe the objective is fully achieved and there is no need to furthur explore the website. Indicate the reason why you think the task objective has been completed within the brackets. E.g., \verb|`|stop [The review and rating information of all the products under electronic category has been tracked. There are 5 pages of products in total and all of them have been visited.]\verb|`|
\end{tcolorbox}
\end{tcolorbox}

\subsection{Implementation Details}
\label{app: details}
\subsubsection{Experiement Details}
In our main experiments, we utilize GPT-4o, GPT-5, GLM-4.5-air-fp8 as backbone models. For GPT-4o and GLM model, following AgentOccam, we utilize the same configuration, setting temperature as 0.5, top\_p as 0.95. For GPT-5, we set reasoning effort to minimal, due to time and budget constraints.

We report results on four domains. Although the WebArena environment also contains a \textit{Map} domain, we found that the service for this website was no longer accessible and therefore excluded it from evaluation. Moreover, since many multi-domain tasks involve the Map website, we also removed these tasks to ensure fair comparison with other methods that reported results only on the remaining domains. 

We also did not compare with AgentSymbiotic~\citep{zhang2025symbiotic} and Learn-by-Interact~\citep{su2025learn}, as the performance of these methods depends heavily on their proprietary retrieval-augmented generation (RAG) databases. Because neither of these works has released their databases, a direct comparison would not be fair or reproducible, and we therefore exclude them from our evaluation.

\subsubsection{Navigation \& Execution}

In our implementation, we follow the action selection mechanism introduced by AgentOccam~\citep{yang2024agentoccam}. Specifically, after the navigation agent generates candidate actions at each step (e.g., clicking an element, entering text, following a link, or stopping), we invoke a separate judge module to evaluate these candidates. The judge receives as input the task instruction, the current observation, the interaction history, and the candidate actions with their rationales. It then ranks or filters the candidates, selecting the action that is most consistent with the high-level objective. 

This design allows the system to correct potential errors from the navigation agent. The judge therefore serves as a lightweight second-opinion layer, ensuring that the final action executed at each step is both safe and aligned with task goals.

During the final execution, if the task requires the analysis result as output, we directly output the analysis result. When writing the analysis code, if there is an error of executing the code, the agent will incorporate the error information and previous code to refine its response to generate another response.
In the other case where the analysis results will be further used to complete web operations (\emph{e.g.,} post a submission in Reddit), {\webweaver} will follow a similar mechanism as navigation, but with the analysis result in the context.

\end{document}

%% file: math_commands.tex
\usepackage{amsmath,amsfonts,bm}

\def\eqref#1{equation~\ref{#1}}

\def\1{\bm{1}}

\DeclareMathAlphabet{\mathsfit}{\encodingdefault}{\sfdefault}{m}{sl}
\SetMathAlphabet{\mathsfit}{bold}{\encodingdefault}{\sfdefault}{bx}{n}



%% file: tables/main.tex
\begin{table}[t]
  \centering
  \small
  \setlength{\tabcolsep}{8pt}
  \renewcommand{\arraystretch}{1.2}
    \caption{Results on the \textbf{WebChoreArena} benchmark across different web domains (Shopping, Reddit, Admin, GitLab). {\webweaver} consistently outperforms all baselines across models , achieving the highest overall success rate. Results with $\dagger$ are reported by WebChoreArena~\citep{miyai2025webchorearena}.} 
    \vspace{2mm}
  \resizebox{\linewidth}{!}{
  \begin{tabular}{l|l|cccc|c}
    \Xhline{1.2pt}
    \rowcolor{CadetBlue!20}
    \textbf{Model} & \textbf{Method} & \textbf{Shopping} & \textbf{Reddit} & \textbf{Admin} & \textbf{GitLab} & \textbf{Overall} \\
    \Xhline{1.2pt}

    \multirow{5}{*}{GPT-5}
      & SteP~\citep{sodhi2023step} &2.6      &4.4      &0.7      &4.7      &3.1  \\
      & BrowserGym~\citep{chezelles2024browsergym} &15.4      &\second{15.4}      & 26.5     &\second{27.6}      &21.2  \\
      & AWM~\citep{wang2024agent} &18.0     & 14.3      &30.3      &26.8      & \second{22.4} \\
      & AgentOccam~\citep{yang2024agentoccam} & \second{21.3} & 11.0 & \second{31.1} & 22.8 &21.6  \\
    \rowcolor{gray!10}
      & {\webweaver}       & \best{35.0}\red{13.7} & \best{26.4}\red{10.0} & \best{34.1}\red{3.0} & \best{29.1}\red{1.5} &\best{31.2}\red{8.8}  \\
    \midrule
    \multirow{5}{*}{GPT-4o}
      & SteP~\citep{sodhi2023step} &2.6      &   0.0   &   0.0  & 4.7     &1.8  \\
      & BrowserGym$\dagger$~\citep{chezelles2024browsergym} & 0.9      & 5.5      & 2.3     & 3.9     &3.2  \\
      & AWM~\citep{wang2024agent} & 3.4     & 8.8     &\second{4.5}     &4.7      &5.4  \\
      & AgentOccam$\dagger$~\citep{yang2024agentoccam} & \second{10.3} & \second{9.9} & \second{4.5}  & \second{7.1}  & \second{8.0}  \\
    \rowcolor{gray!10}
      & {\webweaver}       & \best{18.8}\red{8.5} & \best{19.8}\red{9.9} & \best{12.9}\red{8.4} & \best{9.4}\red{2.3} & \best{15.2}\red{7.2}  \\
    \midrule
    \multirow{5}{*}{GLM-4.5-air-fp8}
      & SteP~\citep{sodhi2023step} & 0.0      &2.2      & 1.5      &2.4      &1.5  \\
      & BrowserGym~\citep{chezelles2024browsergym} &6.0      &4.8      &6.1      & \second{9.4}      &6.6  \\
      & AWM~\citep{wang2024agent} &0.9      &\second{5.6}      &4.3      & 8.7     &4.9  \\
      & AgentOccam~\citep{yang2024agentoccam} & \second{18.8} & 4.4  & \second{11.4} & 8.7  &\second{10.8}  \\
    \rowcolor{gray!10}
      & {\webweaver}       & \best{26.5}\red{7.7} & \best{16.5}\red{10.9} & \best{18.9}\red{7.5} & \best{15.4}\red{6.0} &
      \best{19.3}\red{8.5} \\
    \Xhline{1.2pt}
  \end{tabular}
  }
  \label{tab: main}
\end{table}

%% file: tables/efficiency.tex
\begin{table}[t]
  \centering
  \small
  \setlength{\tabcolsep}{8pt}
  \renewcommand{\arraystretch}{1.2}
    \caption{Efficiency evaluation of \textbf{dynamic re-planning} on WebChoreArena with GPT-4o as backbone LLM. We report accuracy and average navigation steps.}
    \vspace{2mm}
  \resizebox{\linewidth}{!}{
  \begin{tabular}{l|cc|cc|cc|cc}
    \Xhline{1.2pt}
    \rowcolor{CadetBlue!20}
     & \multicolumn{2}{c|}{\textbf{Shopping}} & \multicolumn{2}{c|}{\textbf{Reddit}} & \multicolumn{2}{c|}{\textbf{Admin}} & \multicolumn{2}{c}{\textbf{GitLab}} \\
    \rowcolor{CadetBlue!20}
     & Accuracy & Avg. Steps & Accuracy & Avg. Steps & Accuracy & Avg. Steps & Accuracy & Avg. Steps \\
    \Xhline{1.2pt}
     {\webweaver} & 18.8 & 32.9 & 19.8 & 25.1 & 12.9 & 16.7 & 9.4 & 23.3 \\
     + Dynamic Re-planning.     & 26.5\red{7.7} & 18.2\blue{14.7} & 20.9\red{1.1} & 21.1\blue{4.0} &  13.6\red{0.7} & 17.7\red{1.0} &  11.1\red{1.7} & 21.2\blue{2.1} \\
    \Xhline{1.2pt}
  \end{tabular}
  }
  \label{tab:efficiency}
\end{table}

%% file: tables/webarena.tex
\begin{table}[ht]
  \centering
  \small
  \setlength{\tabcolsep}{8pt}
  \renewcommand{\arraystretch}{1.2}
    \caption{Results on the \textbf{WebArena} benchmark. Bold numbers indicate the best performance, and underlined numbers indicate the second best. All the methods are tested using GPT-4o as backbone model. The baseline results are taken from previous works~\citep{zhang2025webpilot,song2024beyond}.}
    \vspace{2mm}
  \resizebox{0.8\linewidth}{!}{
  \begin{tabular}{l|cccc|c}
    \Xhline{1.2pt}
    \rowcolor{CadetBlue!20}
    \textbf{Method} & \textbf{Shopping} & \textbf{Admin} & \textbf{Reddit} & \textbf{GitLab} & \textbf{Overall} \\
    \Xhline{1.2pt}

    WebArena~\citep{zhou2023webarena}     & 13.9 & 10.4 & 6.6  & 15.0   & 11.5 \\
    AutoEval~\citep{pan2024autonomous}     & \best{39.6} & 20.9 & 20.8 & 25.0   & 26.6 \\
    AWM~\citep{wang2024agent}          & 32.1 & 29.1 & 54.7 & 35.0   & 37.7 \\
    SteP~\citep{sodhi2023step}         & 36.9 & 24.2 & 59.4 & 31.7 & 38.0 \\
    HybridAgent~\citep{song2024beyond}  & 25.7 & \second{41.2} & 51.9 & \second{44.4} & 40.8 \\
    WebPilot~\citep{zhang2025webpilot} & 36.9 & 24.7 & 65.1 & 39.4 & 41.5 \\
    AgentOccam~\citep{yang2024agentoccam}   & \second{37.4} & \best{44.0}   & \second{66.0}   & 38.9 & \second{46.6} \\
    \webweaver    & 36.0   & \second{41.2} & \best{67.9} & \best{47.2} & \best{48.1}  \\
    \Xhline{1.2pt}
  \end{tabular}
  }
  \label{tab:webarena}
\end{table}

%% file: tables/case.tex
\begin{table}[ht]
\centering
\small
\renewcommand{\arraystretch}{1.2}
\caption{Case studies of dynamic re-planning in {\webweaver}.}
\begin{tabularx}{\linewidth}{
  >{\raggedright\arraybackslash}p{0.18\linewidth} |
  >{\raggedright\arraybackslash}p{0.24\linewidth} |
  >{\raggedright\arraybackslash}p{0.22\linewidth} |
  >{\raggedright\arraybackslash}p{0.235\linewidth}}
\Xhline{1.2pt}
\rowcolor{CadetBlue!20}
\makecell[c]{\hdrstrut\textbf{Original Task}} &
\makecell[c]{\hdrstrut\textcolor{RedOrange}{\textbf{Initial Navigation}}\\
                   \textcolor{RedOrange}{\textbf{Objective}}} &
\makecell[c]{\hdrstrut\textbf{Web Elements}\\\textbf{(Description)}} &
\makecell[c]{\hdrstrut\textcolor{GreenPigment}{\textbf{Navigation Objective}}\\
                   \textcolor{GreenPigment}{\textbf{after replanning}}} \\
\Xhline{1.2pt}

Calculate average product price in \textit{Diet \& Sports Nutrition} &
Plan includes navigating to \textit{Diet \& Sports Nutrition} category and going over all the pages. &
Menu to select number of products displayed in each page. &
Add the step changing the number of products displayed each page from 12 to 36. \\

\midrule

Count submissions by specific user \textit{thebelsnickle1991} in each forum &
Decomposition requires traversing submissions in every forum alphabetically, leading to endless exploration. &
Button to submission listing page under the user profile page. &
Revise plan to extract directly from the profile page and aggregate submissions. \\

\midrule

Count unique users among top 600 hottest submissions in \textit{nyc} forum &
Initial plan relies on keyword search for “nyc,” which returns unrelated articles. &
Direct link to the \textit{nyc} forum and its sorting options. &
Bypass search results and directly navigate to the forum page before collecting data. \\

\Xhline{1.2pt}
\end{tabularx}
\label{tab:replanning-cases}
\end{table}

%% file: iclr2026_conference.bbl
\begin{thebibliography}{25}
\providecommand{\natexlab}[1]{#1}
\providecommand{\url}[1]{\texttt{#1}}
\expandafter\ifx\csname urlstyle\endcsname\relax
  \providecommand{\doi}[1]{doi: #1}\else
  \providecommand{\doi}{doi: \begingroup \urlstyle{rm}\Url}\fi

\bibitem[Amayuelas et~al.(2025)Amayuelas, Yang, Agashe, Nagarajan, Antoniades, Wang, and Wang]{amayuelas2025self}
Alfonso Amayuelas, Jingbo Yang, Saaket Agashe, Ashwin Nagarajan, Antonis Antoniades, Xin~Eric Wang, and William Wang.
\newblock Self-resource allocation in multi-agent llm systems.
\newblock \emph{arXiv preprint arXiv:2504.02051}, 2025.

\bibitem[Chezelles et~al.(2024)Chezelles, Le~Sellier, Shayegan, Jang, L{\`u}, Yoran, Kong, Xu, Reddy, Cappart, et~al.]{chezelles2024browsergym}
De~Chezelles, Thibault Le~Sellier, Sahar~Omidi Shayegan, Lawrence~Keunho Jang, Xing~Han L{\`u}, Ori Yoran, Dehan Kong, Frank~F Xu, Siva Reddy, Quentin Cappart, et~al.
\newblock The browsergym ecosystem for web agent research.
\newblock \emph{arXiv preprint arXiv:2412.05467}, 2024.

\bibitem[Drouin et~al.(2024)Drouin, Gasse, Caccia, Laradji, Del~Verme, Marty, Vazquez, Chapados, and Lacoste]{drouin2024workarena}
Alexandre Drouin, Maxime Gasse, Massimo Caccia, Issam~H Laradji, Manuel Del~Verme, Tom Marty, David Vazquez, Nicolas Chapados, and Alexandre Lacoste.
\newblock Workarena: How capable are web agents at solving common knowledge work tasks?
\newblock In \emph{International Conference on Machine Learning}, pp.\  11642--11662. PMLR, 2024.

\bibitem[He et~al.(2024)He, Yao, Ma, Yu, Dai, Zhang, Lan, and Yu]{he2024webvoyager}
Hongliang He, Wenlin Yao, Kaixin Ma, Wenhao Yu, Yong Dai, Hongming Zhang, Zhenzhong Lan, and Dong Yu.
\newblock Webvoyager: Building an end-to-end web agent with large multimodal models.
\newblock In \emph{Proceedings of the 62nd Annual Meeting of the Association for Computational Linguistics (Volume 1: Long Papers)}, pp.\  6864--6890, 2024.

\bibitem[Hong et~al.(2024)Hong, Wang, Lv, Xu, Yu, Ji, Wang, Wang, Dong, Ding, et~al.]{hong2024cogagent}
Wenyi Hong, Weihan Wang, Qingsong Lv, Jiazheng Xu, Wenmeng Yu, Junhui Ji, Yan Wang, Zihan Wang, Yuxiao Dong, Ming Ding, et~al.
\newblock Cogagent: A visual language model for gui agents.
\newblock In \emph{Proceedings of the IEEE/CVF Conference on Computer Vision and Pattern Recognition}, pp.\  14281--14290, 2024.

\bibitem[Koh et~al.(2024)Koh, Lo, Jang, Duvvur, Lim, Huang, Neubig, Zhou, Salakhutdinov, and Fried]{koh2024visualwebarena}
Jing~Yu Koh, Robert Lo, Lawrence Jang, Vikram Duvvur, Ming~Chong Lim, Po-Yu Huang, Graham Neubig, Shuyan Zhou, Ruslan Salakhutdinov, and Daniel Fried.
\newblock Visualwebarena: Evaluating multimodal agents on realistic visual web tasks.
\newblock \emph{arXiv preprint arXiv:2401.13649}, 2024.

\bibitem[Li et~al.(2020)Li, Pinto, and Abbeel]{li2020generalized}
Alexander Li, Lerrel Pinto, and Pieter Abbeel.
\newblock Generalized hindsight for reinforcement learning.
\newblock \emph{Advances in neural information processing systems}, 33:\penalty0 7754--7767, 2020.

\bibitem[Liu et~al.(2018)Liu, Guu, Pasupat, Shi, and Liang]{liu2018reinforcement}
Evan~Zheran Liu, Kelvin Guu, Panupong Pasupat, Tianlin Shi, and Percy Liang.
\newblock Reinforcement learning on web interfaces using workflow-guided exploration.
\newblock \emph{arXiv preprint arXiv:1802.08802}, 2018.

\bibitem[Mialon et~al.(2023)Mialon, Fourrier, Wolf, LeCun, and Scialom]{mialon2023gaia}
Gr{\'e}goire Mialon, Cl{\'e}mentine Fourrier, Thomas Wolf, Yann LeCun, and Thomas Scialom.
\newblock Gaia: a benchmark for general ai assistants.
\newblock In \emph{The Twelfth International Conference on Learning Representations}, 2023.

\bibitem[Miyai et~al.(2025)Miyai, Zhao, Egashira, Sato, Sunada, Onohara, Yamanishi, Toyooka, Nishina, Maeda, et~al.]{miyai2025webchorearena}
Atsuyuki Miyai, Zaiying Zhao, Kazuki Egashira, Atsuki Sato, Tatsumi Sunada, Shota Onohara, Hiromasa Yamanishi, Mashiro Toyooka, Kunato Nishina, Ryoma Maeda, et~al.
\newblock Webchorearena: Evaluating web browsing agents on realistic tedious web tasks.
\newblock \emph{arXiv preprint arXiv:2506.01952}, 2025.

\bibitem[Pan et~al.(2024)Pan, Zhang, Tomlin, Zhou, Levine, and Suhr]{pan2024autonomous}
Jiayi Pan, Yichi Zhang, Nicholas Tomlin, Yifei Zhou, Sergey Levine, and Alane Suhr.
\newblock Autonomous evaluation and refinement of digital agents.
\newblock \emph{arXiv preprint arXiv:2404.06474}, 2024.

\bibitem[Shi et~al.(2017)Shi, Karpathy, Fan, Hernandez, and Liang]{shi2017world}
Tianlin Shi, Andrej Karpathy, Linxi Fan, Jonathan Hernandez, and Percy Liang.
\newblock World of bits: An open-domain platform for web-based agents.
\newblock In \emph{International Conference on Machine Learning}, pp.\  3135--3144. PMLR, 2017.

\bibitem[Shinn et~al.(2023)Shinn, Cassano, Gopinath, Narasimhan, and Yao]{shinn2023reflexion}
Noah Shinn, Federico Cassano, Ashwin Gopinath, Karthik Narasimhan, and Shunyu Yao.
\newblock Reflexion: Language agents with verbal reinforcement learning.
\newblock \emph{Advances in Neural Information Processing Systems}, 36:\penalty0 8634--8652, 2023.

\bibitem[Sodhi et~al.(2023)Sodhi, Branavan, Artzi, and McDonald]{sodhi2023step}
Paloma Sodhi, SRK Branavan, Yoav Artzi, and Ryan McDonald.
\newblock Step: Stacked llm policies for web actions.
\newblock \emph{arXiv preprint arXiv:2310.03720}, 2023.

\bibitem[Song et~al.(2024)Song, Xu, Zhou, and Neubig]{song2024beyond}
Yueqi Song, Frank Xu, Shuyan Zhou, and Graham Neubig.
\newblock Beyond browsing: Api-based web agents.
\newblock \emph{arXiv preprint arXiv:2410.16464}, 2024.

\bibitem[Su et~al.(2025)Su, Sun, Yoon, Yin, Yu, and Ar{\i}k]{su2025learn}
Hongjin Su, Ruoxi Sun, Jinsung Yoon, Pengcheng Yin, Tao Yu, and Sercan~{\"O} Ar{\i}k.
\newblock Learn-by-interact: A data-centric framework for self-adaptive agents in realistic environments.
\newblock \emph{arXiv preprint arXiv:2501.10893}, 2025.

\bibitem[Wang et~al.(2024)Wang, Mao, Fried, and Neubig]{wang2024agent}
Zora~Zhiruo Wang, Jiayuan Mao, Daniel Fried, and Graham Neubig.
\newblock Agent workflow memory.
\newblock \emph{arXiv preprint arXiv:2409.07429}, 2024.

\bibitem[Wei et~al.(2025)Wei, Sun, Papay, McKinney, Han, Fulford, Chung, Passos, Fedus, and Glaese]{wei2025browsecomp}
Jason Wei, Zhiqing Sun, Spencer Papay, Scott McKinney, Jeffrey Han, Isa Fulford, Hyung~Won Chung, Alex~Tachard Passos, William Fedus, and Amelia Glaese.
\newblock Browsecomp: A simple yet challenging benchmark for browsing agents.
\newblock \emph{arXiv preprint arXiv:2504.12516}, 2025.

\bibitem[Yang et~al.(2024{\natexlab{a}})Yang, Liu, Chaudhary, Fakoor, Chaudhari, Karypis, and Rangwala]{yang2024agentoccam}
Ke~Yang, Yao Liu, Sapana Chaudhary, Rasool Fakoor, Pratik Chaudhari, George Karypis, and Huzefa Rangwala.
\newblock Agentoccam: A simple yet strong baseline for llm-based web agents.
\newblock \emph{arXiv preprint arXiv:2410.13825}, 2024{\natexlab{a}}.

\bibitem[Yang et~al.(2025)Yang, Li, Dai, Yang, Luo, Zhao, Hu, Huang, Saha, Chen, et~al.]{yang2025gta1}
Yan Yang, Dongxu Li, Yutong Dai, Yuhao Yang, Ziyang Luo, Zirui Zhao, Zhiyuan Hu, Junzhe Huang, Amrita Saha, Zeyuan Chen, et~al.
\newblock Gta1: Gui test-time scaling agent.
\newblock \emph{arXiv preprint arXiv:2507.05791}, 2025.

\bibitem[Yang et~al.(2024{\natexlab{b}})Yang, Wang, Li, Luo, Chen, Huang, and Li]{yang2024aria}
Yuhao Yang, Yue Wang, Dongxu Li, Ziyang Luo, Bei Chen, Chao Huang, and Junnan Li.
\newblock Aria-ui: Visual grounding for gui instructions.
\newblock \emph{arXiv preprint arXiv:2412.16256}, 2024{\natexlab{b}}.

\bibitem[Yao et~al.(2023)Yao, Zhao, Yu, Du, Shafran, Narasimhan, and Cao]{yao2023react}
Shunyu Yao, Jeffrey Zhao, Dian Yu, Nan Du, Izhak Shafran, Karthik Narasimhan, and Yuan Cao.
\newblock React: Synergizing reasoning and acting in language models.
\newblock In \emph{International Conference on Learning Representations (ICLR)}, 2023.

\bibitem[Zhang et~al.(2025{\natexlab{a}})Zhang, Qiu, Tan, Zhang, Lu, Peng, Xu, Agudelo, Qian, and Chen]{zhang2025symbiotic}
Ruichen Zhang, Mufan Qiu, Zhen Tan, Mohan Zhang, Vincent Lu, Jie Peng, Kaidi Xu, Leandro~Z Agudelo, Peter Qian, and Tianlong Chen.
\newblock Symbiotic cooperation for web agents: Harnessing complementary strengths of large and small llms.
\newblock \emph{arXiv preprint arXiv:2502.07942}, 2025{\natexlab{a}}.

\bibitem[Zhang et~al.(2025{\natexlab{b}})Zhang, Ma, Ma, Han, Wu, and Tresp]{zhang2025webpilot}
Yao Zhang, Zijian Ma, Yunpu Ma, Zhen Han, Yu~Wu, and Volker Tresp.
\newblock Webpilot: A versatile and autonomous multi-agent system for web task execution with strategic exploration.
\newblock In \emph{Proceedings of the AAAI Conference on Artificial Intelligence}, volume~39, pp.\  23378--23386, 2025{\natexlab{b}}.

\bibitem[Zhou et~al.(2023)Zhou, Xu, Zhu, Zhou, Lo, Sridhar, Cheng, Ou, Bisk, Fried, et~al.]{zhou2023webarena}
Shuyan Zhou, Frank~F Xu, Hao Zhu, Xuhui Zhou, Robert Lo, Abishek Sridhar, Xianyi Cheng, Tianyue Ou, Yonatan Bisk, Daniel Fried, et~al.
\newblock Webarena: A realistic web environment for building autonomous agents.
\newblock \emph{arXiv preprint arXiv:2307.13854}, 2023.

\end{thebibliography}
